\documentclass[conference]{IEEEtran}
\IEEEoverridecommandlockouts
\usepackage{cite}
\usepackage{amsmath,amssymb,amsfonts}
\usepackage{algorithmic}
\usepackage{graphicx}
\usepackage{textcomp}
\usepackage{xcolor}
\def\BibTeX{{\rm B\kern-.05em{\sc i\kern-.025em b}\kern-.08em
    T\kern-.1667em\lower.7ex\hbox{E}\kern-.125emX}}

\usepackage{cite}
\usepackage{amsmath,amssymb,amsfonts}
\usepackage{algorithmic}
\usepackage{graphicx}
\usepackage{textcomp}
\usepackage{xcolor}
\usepackage[utf8]{inputenc}
\usepackage{caption}
\usepackage{amssymb}
\usepackage{mathtools}
\usepackage{todonotes}
\usepackage{balance}

\usepackage{booktabs}
\usepackage{multirow}
\usepackage{adjustbox}
\usepackage{colortbl}
\usepackage{tikz}
\usepackage{collcell}

\usepackage[hyperfootnotes=false]{hyperref}

\begin{document}

\title{PULL: Reactive Log Anomaly Detection\\Based On Iterative PU Learning}

\author{\IEEEauthorblockN{Thorsten Wittkopp, Dominik Scheinert, Philipp Wiesner, Alexander Acker, and Odej Kao}
\IEEEauthorblockA{\textit{Technische Universität Berlin, Germany} \\
\{t.wittkopp, dominik.scheinert, wiesner, alexander.acker, odej.kao\}@tu-berlin.de}
}

\maketitle

\begin{abstract}
Due to the complexity of modern IT services, failures can be manifold, occur at any stage, and are hard to detect.
For this reason, anomaly detection applied to monitoring data such as logs allows gaining relevant insights to improve IT services steadily and eradicate failures.
However, existing anomaly detection methods that provide high accuracy often rely on labeled training data, which are time-consuming to obtain in practice.
Therefore, we propose PULL, an iterative log analysis method for reactive anomaly detection based on estimated failure time windows provided by monitoring systems instead of labeled data.
Our attention-based model uses a novel objective function for weak supervision deep learning that accounts for imbalanced data and applies an iterative learning strategy for positive and unknown samples (PU learning) to identify anomalous logs.
Our evaluation shows that PULL consistently outperforms ten benchmark baselines across three different datasets and detects anomalous log messages with an F1-score of more than 0.99 even within imprecise failure time windows.
\end{abstract}

\section{Introduction}
\label{sec:introduction}

Modern IT services are increasingly complex, which aggravates their operation and maintenance and poses new challenges for operators~\cite{rosendo2018improve}.
Monitoring solutions are commonly employed to observe and optimize IT services (e.g. to comply with service level agreements~(SLA)). 
However, the amount of monitoring data oftentimes hinders manual analysis, which demands automated approaches.
To account for that, the emerging field of artificial intelligence for IT operations (AIOps) intends to support engineers~\cite{gulenko2020ai}.
They can improve reliability and stability in further service updates through detected failures in IT services by AIOps. 
For this, the core components of any AIOps system are the three pillars of observability, namely metrics, traces, and logs~\cite{sridharan2018distributed}.
The latter are important resources for troubleshooting because they record events during the execution of service applications~\cite{wittkopp2021taxonomy}.
Although most log messages come with a severity level, such as INFO, WARNING, and ERROR, it does not necessarily reflect the overall status and oftentimes exhibits imprecision.
Hence, recent research on anomaly detection is based on deep learning (DL) models to analyze log messages~\cite{du2017deeplog,zhang2019robust,Wittkopp_LogLab_2021,nedelkoski2020self,wittkopp2021a2log}.

While the majority of existing anomaly detection methods focus on predictive - live - anomaly detection, these methods are often not directly applicable in real-world services due to insufficient performance~\cite{yang2021semi} or the explicit need for accurately labeled data, which is costly and time-consuming to obtain~\cite{wen2021time}. 
These limitations can be mitigated through analysis in retrospect - reactive - as more information becomes available.
Especially IT companies, which operate complex services and frequently deploy new versions, need to assume that the heterogeneity of users and devices reveals not yet observed errors over time. 
Therefore, they have an interest in thoroughly investigating all occurred anomalies to properly update and steadily improve their services.
Reactive anomaly detection qualifies especially for reliability engineering, which operates in longer iterations and attempts to maximize long-term functionality.

\begin{figure*}[htbp]
\centering
\includegraphics[width=0.95\textwidth]{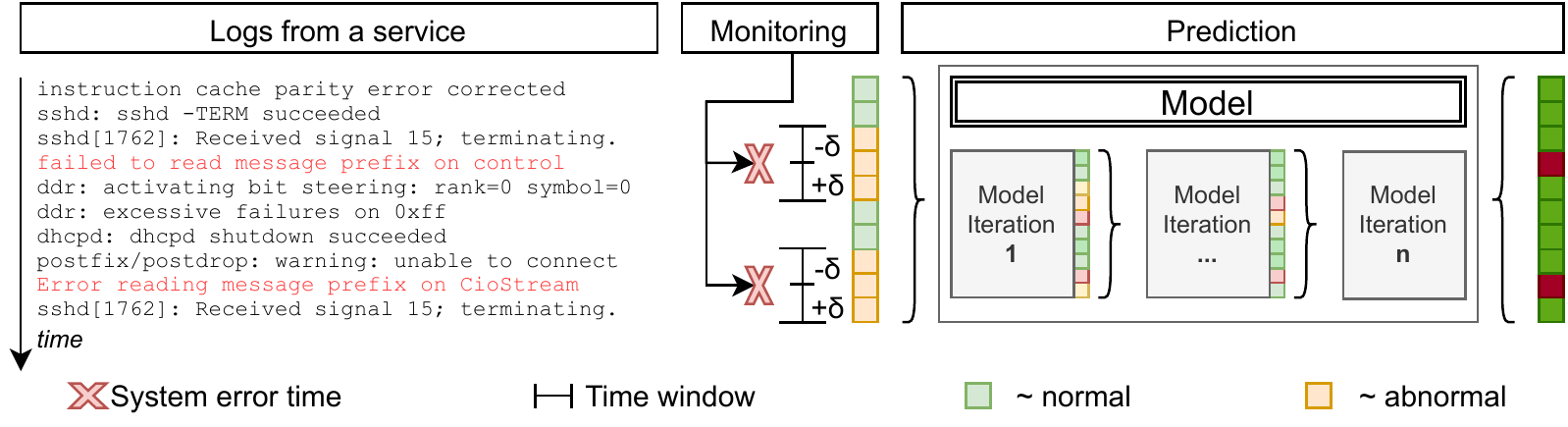}
\caption{We use rough estimates for failure times provided by monitoring systems to identify abnormal log messages via weak supervision.}
\label{fig:problem_description}
\end{figure*}

We propose PULL, an attention-based model for reactive log anomaly detection.
Instead of labeled data, it relies on rough estimates of when an error has occurred - information that can often be derived from monitoring systems~\cite{sukhwani2017monitoring}.
PULL utilizes PU learning~\cite{liu2002partially,liu2003building}, a weak supervision technique, to identify abnormal log messages in the estimated failure time windows and therefore provides substantial insights to the reliability engineer. 
Our method benefits from a novel iterative training strategy where a new model is instantiated after every iteration. It receives the results of the previous model as input.
Specifically, the contributions of this paper are:

\begin{itemize}
    \item A new method for reactive anomaly detection called PULL, which is based on the transformer architecture~\cite{DevlinCLT19} with attention mechanism~\cite{VaswaniSPUJGKP17}. 
    \item An objective function for weak supervision deep learning that takes class-imbalanced data into account and is applicable to our iterative learning strategy.
    \item An evaluation of PULL against ten baselines (DeepLog, LogRobust, Rocchio, Invariants Miner, SVM, boosting methods, PCA, Logistic Regression, Decision Tree, and Random Forest) regarding solving the defined problem and their applicability for our proposed iterative training strategy. 
\end{itemize}

The remainder of this paper is structured as follows.
\autoref{sec:towards} provides background and formally describes our problem.
\autoref{sec:related-work} surveys the related work.
\autoref{sec:approach} explains our method in detail and elaborates on its advantages.
\autoref{sec:evaluation} evaluates PULL in comparison to ten other methods.
\autoref{sec:conclusion} concludes the paper.

\section{Towards Reactive Anomaly Detection}
\label{sec:towards}

Log events record the execution path of an IT service.
They result from logging instructions that are part of the source code.
Log events can describe failures that occur during runtime, such as the crash of a service.
We call such log events abnormal.
Log events are marked with a severity level, that indicates how serious a log event is regarding a potential failure.
However, even logs that are not marked with high severity can be important indicators for faults and important for the reliability engineers.
Other log events may be assigned a high severity, but the event has no adverse effects on the service.
Therefore, by only looking at the severity of a log event, it can hardly be concluded if a log message needs to be considered abnormal or not.
This results in high efforts for the engineers to find the real abnormal log events by hand in order to investigate the causes of failures.

Modern monitoring solutions alert when a service operates outside of a defined norm.
They observe system metrics of hardware and software components. Even if most failures can be classified, not all failures can be known in advance~\cite{sukhwani2017monitoring}.
Nevertheless, we assume that failure times are roughly known, through the monitoring solution.
We use this information in retrospect to detect abnormal logs on the log message level in an automated way.
Therefore we use the failure time windows from the monitoring system as an auxiliary source to train our method without previously known labels.

\autoref{fig:problem_description} exemplifies the described problem.
It displays the log of a system with two abnormal log events colored in red for illustration purposes.
As described above, we utilize monitoring information to estimate time windows of the length $2*\delta$ in which we suspect abnormal log events to be present. 
The goal of the method is to identify these abnormal log messages and classify all others as normal, regardless of whether they 
are included in the respective time windows.

Our reactive anomaly detection method has the main advantage that it enables us to determine the failures with a high degree of precision and recall, without the need for labeled data and thus reducing the engineer's workload in finding failures to improve the service.

\textbf{Problem Definition.}
We describe our log anomaly detection task as a weak supervision learning problem with inaccurate labels since we cannot assign accurate labels due to imprecision of the failure time windows. Weak supervision with inaccurate labels is defined as a situation where the supervision information is not always matching the ground-truth~\cite{zhou2018brief}.
Therefore, we assign preliminary abnormal labels for all log events in the estimated time windows and preliminary normal labels for all other log events. 
We utilize PU learning~\cite{liu2002partially,liu2003building} which is short for learning from positive and unlabeled data.
PU learning is an umbrella term for several weakly supervised binary classification methods that classify unlabeled samples by learning the positive (normal) and treating the unlabeled as abnormal \cite{liu2002partially,zhu2009introduction,bekker2020learning}. 
For this work, we assume that \raisebox{.5pt}{\textcircled{\raisebox{-.9pt} {1}}} the approximated normal class mainly consists of true normal samples, and \raisebox{.5pt}{\textcircled{\raisebox{-.9pt} {2}}} the characteristics of the real abnormal samples, hidden in the unlabeled samples, differ from those of the normal samples.
Furthermore, we assume that at least 50 \% of the samples are normal.

Log anomaly detection is a binary classification problem for a log dataset $\mathcal{L}$.
Using weak supervision with inaccurate labels, each log event $l_i \in \mathcal{L}$ is assigned a triple $(x_i, \tilde{y}_i, y_i)$, where $x_i$ is the preprocessed log message, $\tilde{y}_i$ the inaccurate label, and $y_i$ the ground truth. The ground truth $y_i$ is only available in the experiment setup, to evaluate and compare different methods. 
We define $\tilde{y}_i, y_i \in \{0,1\}$, where $0$ is the label for normal and $1$ for abnormal log events, and form the two disjunct classes $\mathcal{P} = (x_i\ |\ \forall l_i \in \mathcal{L}, \tilde{y}_i = 0)$ as well as $\mathcal{U} = (x_i\ |\ \forall l_i \in \mathcal{L}, \tilde{y}_i = 1)$.
Let $\Phi(x, \tilde{y}, \Theta) : l_i \rightarrow \{0,1\}$ be a function represented by a trainable model, that is trained on the incoming log messages $x$ by using their inaccurate labels $\tilde{y}$, to learn the model parameters $\Theta$. We then use these parameters to predict a label $\widehat{y}_i = \Phi(x_i, \Theta)$ for each message $x_i$, where $\Phi(x_i, \Theta)$ is the model output.

\section{Related Work}
\label{sec:related-work}
Next, we discuss related works concerning PU learning and log anomaly detection.

\textbf{PU Learning.}
The authors in~\cite{liu2002partially} utilize the Expectation-Maximization (EM) algorithm together with Naive Bayes classification, with the EM algorithm eventually producing a sequence of Naive Bayes classifiers. 
A more conservative variant of this method is proposed in~\cite{fusilier2015detecting} where the set of reliable negative (RN) instances is iteratively pruned using a binary classifier, which ultimately leads to improved final prediction results due to the few but high quality negative instances.
In~\cite{WuCWWZW20}, the authors extract a set of reliable negative instances via a feature strength function using a greedy heuristic on sorted features until $|\mathcal{P}|\simeq|RN|$. 
In another work~\cite{LuoCLJ18}, the set RN is determined using an ensemble strategy that integrates both Naive Bayes and Logistic Regression, where an agreement of both models is required to extend the set. 
An ensemble learning method for PU learning is proposed in~\cite{mordelet2014bagging}. The authors motivate bagging SVM, i.e. the aggregation of multiple SVM classifiers to answer sources of instability often encountered in PU learning.
They find their approach to producing competitive results while being often faster to train, especially for $|\mathcal{P}| \ll |\mathcal{U}|$.

\textbf{Log Anomaly Detection.}
Various methods have been presented for text-based problems, and more specifically, for log anomaly detection. 
Existing literature~\cite{he2016experience,kowsari2019text} commonly clusters them into supervised and unsupervised methods.

\textit{Supervised Methods} assume the presence of labeled data for training.
Logistic regression~\cite{hosmer2013applied} has demonstrated its usefulness for classification tasks~\cite{genkin2007large}.
Decision Trees~\cite{quinlan1986induction} are another solution often employed in classification problem scenarios~\cite{safavian1991survey,chen2004failure}.
Random forests~\cite{ho1995random} build upon decision trees and are a suitable tool for classification due to their ensemble learning design. 
SVMs are evaluated in~\cite{manevitz2001one,liang2007failure} for anomaly detection and are most often among the best performing methods.
Other publications~\cite{sowmya2016large,selvi2017text} utilize the Rocchio algorithm and demonstrate its effectiveness, e.g., for the task of large-scale multi-label text classification.
In recent years, deep learning solutions have been proposed for the problem of supervised anomaly detection on logs.
LogRobust~\cite{zhang2019robust} tackles the instability issue of established methods via an attention-based Bi-LSTM model and semantic vectorization of log events.
This idea is further developed with SwissLog~\cite{LiCJHY20}, which explicitly leverages semantic and temporal information to handle diverse faults. 
LogBERT~\cite{GuoYW21} extends BERT for log anomaly detection via two specifically designed self-supervised training tasks. 
An attention-based encoder model is also used with Logsy~\cite{nedelkoski2020self}, however, additional anomaly samples from auxiliary log datasets are used to enhance the learned vector representations of normal log data.
PLELog~\cite{YangCWWJDZ21} designs an attention-based GRU neural network, uses the idea of PU learning and conducts probabilistic label estimation.

In contrast, \textit{Unsupervised Methods} can be used without labeled data.
The PCA algorithm~\cite{jolliffe2005principal} is employed for dimensionality reduction right before the actual classification procedure. 
Invariant Miners~\cite{lou2010mining} retrieve structured logs using log parsing, group log messages according to log parameter relationships, and mine program invariants from groupings in an automated fashion for log anomaly detection.
The authors of~\cite{schapire2000boostexter} propose a boosting-based ensemble learning method that shows good performance on a standard benchmark problem. 
LogCluster~\cite{LinZLZC16} is a clustering-based method that relies on log vectorization, clustering via Agglomerative Hierarchical Clustering, and extraction of cluster representatives.
Recently, deep learning methods have been also introduced for unsupervised log anomaly detection.
DeepLog~\cite{du2017deeplog} utilizes templates~\cite{he2017drain} and an LSTM, interprets a log as a sequence of sentences, and performs anomaly detection on each.
LogAnomaly~\cite{MengLZZPLCZTSZ19} introduces the representation method template2vec and fuses it with LSTM networks into an end-to-end framework that detects sequential and quantitative anomalies. 
In~\cite{FarzadG20}, the authors propose the combination of Isolation Forests and multiple Autoencoder Networks. 
ADA~\cite{YuanALYL020} employs LSTM networks, dynamic thresholding, and intelligent data storing.

In light of the existing deep learning methods, PULL differentiates from them in multiple aspects.
It applies to individual log messages, hence it is not constrained to log sequences only.
Lastly, our method introduces a novel approach to weak-supervised anomaly detection of log messages via the idea of failure time windows and PU learning through an iterative training strategy.

\section{Method}
\label{sec:approach}

This section introduces our method for log anomaly detection. 
As the absence of accurately labeled data is a common problem in real-world use cases, we design a training pipeline that leverages various techniques and proposes a novel loss function that enables us to detect anomalies in logs, by training on inaccurately labeled data from monitoring solutions.

\subsection{Preliminaries}
To increase the transparency of software systems and to ease problem understanding, logging is commonly employed.
We differentiate the available log data into two classes, namely the positive class $\mathcal{P}$ and the unlabeled class $\mathcal{U}$.
While $\mathcal{U}$ encompasses all log messages that occur in the time windows, $\mathcal{P}$ describes the rest. 
The unlabeled class is called \emph{unknown} since it most likely contains normal and abnormal data, yet its members are at first sight treated as abnormal and thus assigned preliminary abnormal labels.
This is in contrast to standard anomaly detection where the common notation is that the positive class are anomalous log lines. 
Since we are referring to PU learning, we will stick to the notation of PU learning.

Each log instruction results in a single log event, such that the complete log is a sequence of events $\mathcal{L} = (l_i : i = 1,2, \ldots n)$. 
A log event $l_i \in \mathcal{L}$ can be further decomposed into \emph{meta-information} and the \emph{content} $c_i$. In accordance to our problem definition, the \emph{content} $c_i$ serves as the model input $x_i$.

\textbf{Tokenization.} In order to input the content of log messages $c_i$ to an algorithm, an initial transformation with methods from the domain of Natural Language Processing (NLP) is employed. 
Given a log content $c_i$, we refer to the smallest indecomposable unit as a \emph{token}. 
Consequently, each log content $c_i$ can be interpreted as a sequence $c_i = (w_j: w_j \in V, j = 1,2,\ldots, s_i)$ of tokens, where $w_j$ is the $j$-th token in $c_i$, $V$ is a set of all known tokens and thus the used \emph{vocabulary}, and $s_i$ denotes the total number of tokens in $c_i$. 

\textbf{Embeddings.} 
As the tokens themselves are elements of the vocabulary $V$, they can not be passed into a neural network directly. Furthermore, tokens do not provide any information about their similarity or difference to each other, hence, so called \emph{embeddings} are used to compute a representation of the tokens such that a machine learning model can process it.
An embedding $\Vec{v}$ is a real-valued vector representation $\Vec{v} \in \mathbb{R}^d$ of a token; a transformation function $g$ transforms a sequence of tokens $c_i$ with length $s_i$ into a sequence of embeddings $\vec{e_i}$, i.e. with $g: V^{|s_i|} \rightarrow \mathbb{R}^{d,|s_i|}$. 
To access the $j$-th embedding in a sequence of embeddings $\vec{e_i}$, we write $\vec{e_i}(j)$. Furthermore, they are used to obtain a representation for the whole sequence of tokens $c_i$. Embeddings are trainable units adapted during the model training process to, at best, represent the meaning of the original token or sequence of tokens.

\subsection{Anomaly Detection}
\begin{figure}[htbp]
\centering
\includegraphics[width=0.9\columnwidth]{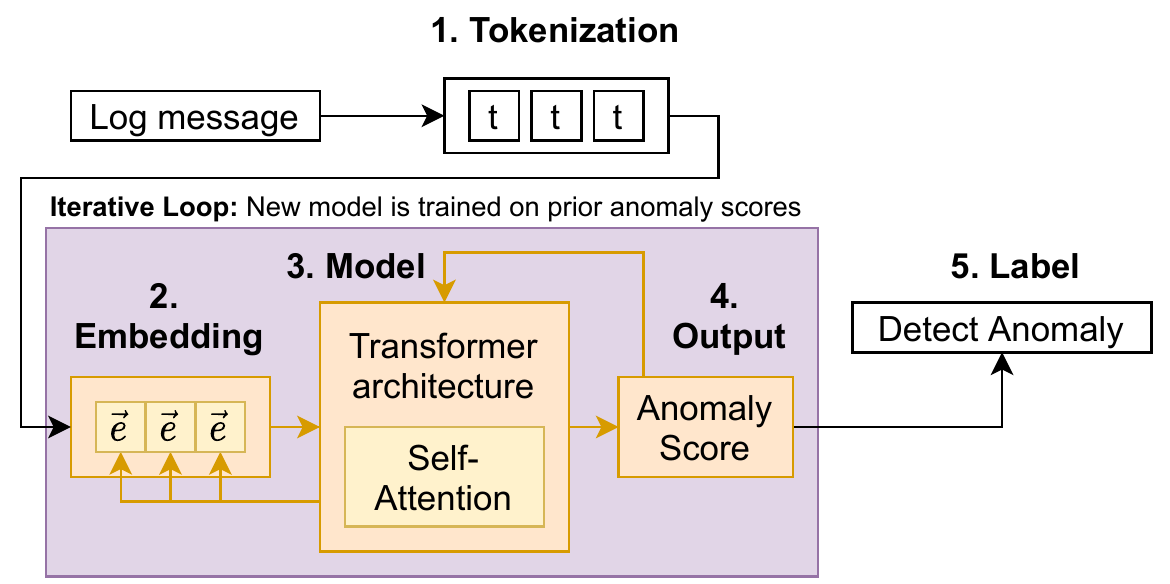}
\caption{High level pipeline}
\label{fig:high_level_pipeline}
\end{figure}

We design a processing pipeline for the anomaly detection in logs, which is illustrated in~\autoref{fig:high_level_pipeline}.

First, we convert the content $c_i$ of each log event $l_i$ into a sequence of tokens $c_i$ using the symbols \texttt{.,:/} and whitespaces as separators. Subsequently, we clean the resulting sequence of tokens by replacing certain tokens with placeholders that adequately represent the original token without losing relevant information. We introduce a placeholder token \texttt{'[HEX]'} for hexadecimal values, as well as \texttt{'[NUM]'} for any number greater or equal 10. Finally, we prefix the sequence of transformed tokens with a special placeholder token \texttt{'[CLS]'} which will be beneficial later on. 
An exemplary log message
\begin{center}
    \footnotesize
    \texttt{time.c: Detected 3591.142 MHz.}
\end{center}
is thus transformed into a sequence of tokens
\begin{center}
    \footnotesize
    \texttt{['[CLS]', 'time', 'c', 'Detected', '[NUM]', '[NUM]', 'MHz']}.
\end{center}

Since these sequences can vary in length, we truncate them to a fixed size $s$ and fill up smaller sequences with padding tokens \texttt{'[PAD]'}.
For each token $w_j$ in the token sequence $c_i$, an embedding vector $\vec{e}_{i}(j)$ is obtained using the transformation function $g$.
The truncated sequences of embeddings $\vec{e_i}'$ serve as input for the model. 
The model computes an output embedding for each truncated input embedding sequence $\vec{e_i}'$, which summarizes the log message by utilizing the embeddings of all tokens. 
This output embedding is encoded in the embedding of the \texttt{’[CLS]’} token and modified during training via loss minimization. 
During the training process, the model is supposed to learn the meanings of the log messages, thereby getting an intuition of what is normal and abnormal, respectively.
We denote the output of the model as $z_i$ and use it throughout the remaining steps.
The anomaly score is calculated by the length of the output vector $\lVert z_i \rVert$.
Anomaly scores close to $0$ represent normal log messages, whereas large scores indicate an abnormal log message. 
Though we conduct an iterative training strategy where the computed anomaly scores are transformed and used as input to subsequent training iterations, after the last iteration, the final computed anomaly score of each log event $l_i$ is eventually used to assign a label $\widehat{y_i}$, i.e. either normal or abnormal, based on a determined threshold.
In the following subsections, we will explain our model architecture, the objective function to train the model, as well as our iterative training strategy in more detail.

\subsection{Model Architecture}
\begin{figure}
    \centering
    \includegraphics[width=0.95\columnwidth]{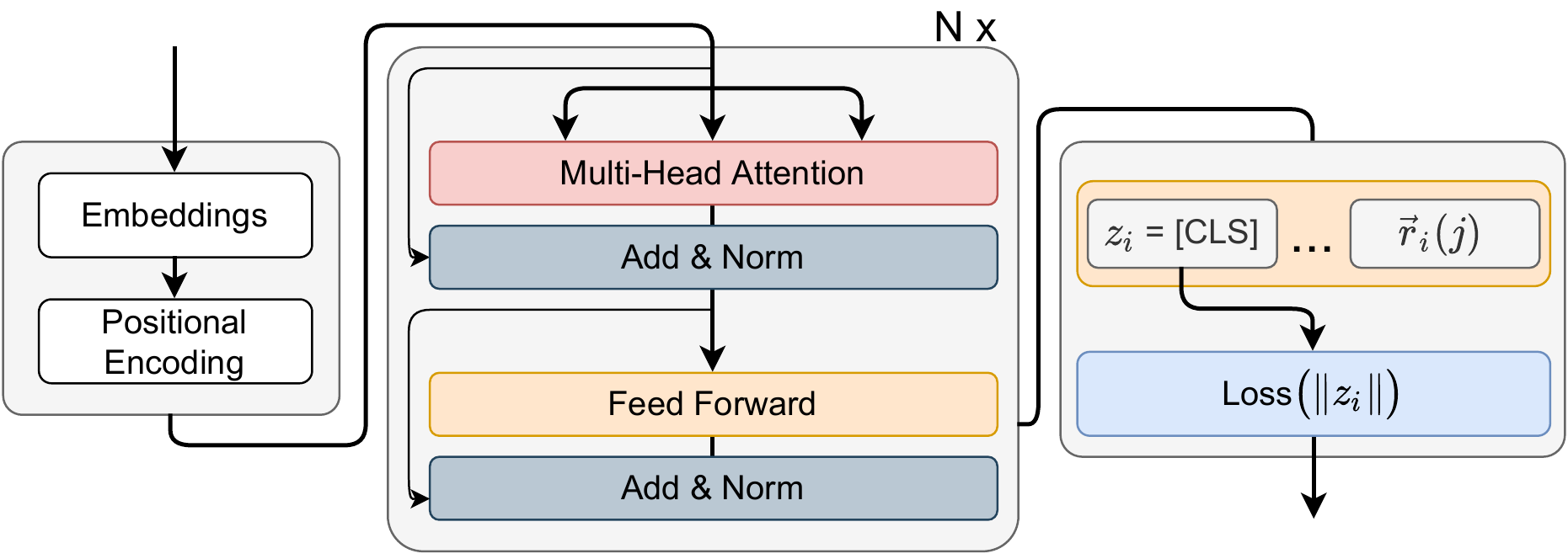}
    \caption{Transformer encoder architecture with multi-head self-attention. We define a novel loss function on the encoder output operating on distances.}
    \label{fig:approach_model}
\end{figure}

For our network architecture illustrated in~\autoref{fig:approach_model}, we utilize the encoder of the transformer architecture~\cite{DevlinCLT19} with self-attention~\cite{VaswaniSPUJGKP17}. Since this architecture does not take the order of the input embeddings into account, we further enrich them with \emph{positional encoding}~\cite{GehringAGYD17,VaswaniSPUJGKP17}.
Through this encoding of the embeddings, the order of tokens is preserved. 

Our employed encoder architecture utilizes the attention mechanisms.
Most importantly, the attention mechanisms allow for attending over input embeddings and thus determining their overall importance, whereas the utilization of multiple attention mechanisms simultaneously (i.e. multi-head) stabilizes the learning process.
Finally, this model outputs a transformed vector representation $\vec{r}_i(j)$ for each input embedding $\vec{e_i}(j)$ representing a source token $w_j$. 
Furthermore, during training, the input embedding corresponding to the \texttt{’[CLS]’} token attends over all other input embeddings from the original sequence of tokens, which enables the model to summarize the context of the log message content $c_i$ of the \texttt{’[CLS]’} token.

As we solely use this vector representation in the subsequent steps, we hence refer to it as $z_i$ instead of $\Phi(\vec{e_i}'; \Theta)$ for the sake of simplicity.

\subsection{Objective Function}
To label log data, the transformer model must be trained in a way that it is capable to handle the problem of weak supervision with inaccurate labels. 
For this, the model must understand the semantics of the log messages. Thereby several log messages can be completely different, but express the same state of the system and have therefore the same meaning. 
Thus, the objective function must be modeled in a way that log events occurring in both $\mathcal{P}$ and $\mathcal{U}$ have low anomaly scores. 
Log events that occur in $\mathcal{U}$ only are most likely abnormal and must therefore induce higher anomaly scores.
In addition, the loss function must be able to handle large amounts of incorrectly labeled log messages, since the class $\mathcal{U}$ can increase quickly for large $\delta$, as $\delta$ is only roughly estimated from the monitoring systems.
The anomaly scores for each input sequence $\vec{e_i}'$ are calculated as the length of the corresponding outcome vector $\lVert z_i\rVert$. 
The length is calculated by the euclidean distance to the zero vector.

The objective function consists of two parts. The first part minimizes the errors of samples from class $\mathcal{P}$, from which the calculated anomaly scores should be small and close to $0$. The second part minimizes the errors of samples from class $\mathcal{U}$, by enlarging the anomaly scores sufficiently, which makes them diverge from $0$. 
The general structure of the objective function is shown in \autoref{eq:objective_function_gen}, where $\tilde{y}_i$ is the inaccurate label, $z_i$ is the output vector representation of the model for each embedded input log message $\vec{e_i}'$, and $m$ the number of samples per batch.

\begin{equation}
    \label{eq:objective_function_gen}
    \frac{1}{m}\sum\limits_{i=1}^{m}((1-\tilde{y}_i)*a(z_i) + (\tilde{y}_i)*b(z_i)
\end{equation}

The first part $a$ of the objective function becomes $0$ if the sample is from class $\mathcal{U}$, while the second part $b$ becomes $0$ if the sample is from class $\mathcal{P}$. For $a$ we minimize the error for positive samples and in contrast, we increase the error for all anomaly scores, when the log message is of class $\mathcal{U}$, with $a(z_i) = \lVert z_i \rVert^2$ and $b(z_i) = q^2 / \lVert z_i \rVert$,
where $q$ is a numerator between 0 and 1 that represents the relation of the number of samples in $\mathcal{P}$ and $\mathcal{U}$.
To ensure that $q$ is representing the relation of $\mathcal{P}$ and $\mathcal{U}$ and remains in the boundaries of 0 to 1 without growing too fast, we model $q$ as a limited function: $q = f(\frac{|\mathcal{P}|}{|\mathcal{U}|}) = \frac{|\mathcal{P}| / |\mathcal{U}|}{(|\mathcal{P}| / |\mathcal{U}|)+1} = \frac{|\mathcal{P}|}{|\mathcal{P}|+|\mathcal{U}|}$, where $f(x) = \frac{x}{x+1}, \lim \limits_{x \to \infty} f(x) = 1$.
The limited function $f$ with limes of 1 enforces the requirements for $q$.
Thus, the total loss function is composed as

\begin{equation}
    \frac{1}{m}\sum\limits_{i=1}^{n}\Big((1-y)*\lVert z_i \rVert^2 + (y)*\frac{(\frac{|\mathcal{P}|}{|\mathcal{P}|+|\mathcal{U}|})^2}{\lVert z_i \rVert}\Big),
\end{equation}

which effectively enables the transformer model to train log messages with inaccurate labels by modifying the calculated error depending on the relation of $\mathcal{P}$ and $\mathcal{U}$.

\subsection{Iterative Training Strategy}

\begin{figure}
    \centering
    \includegraphics[width=0.8\columnwidth]{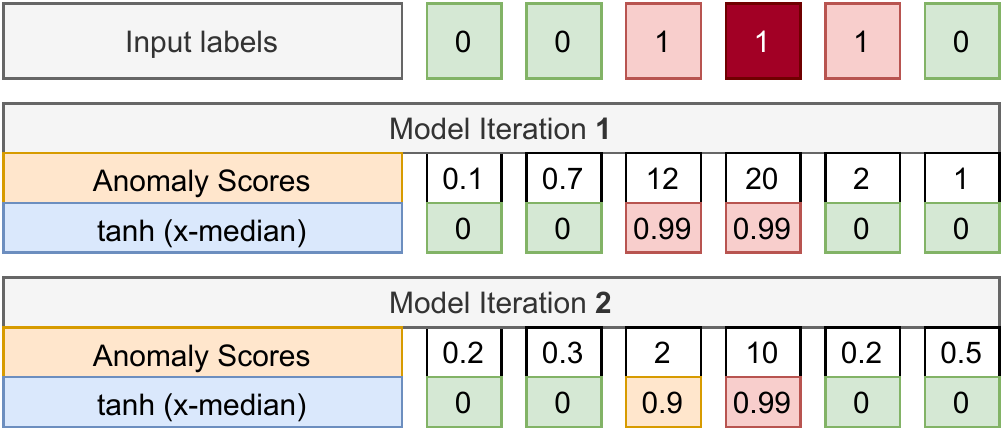}
    \caption{Anomaly scores are smoothed for training.}
    \label{fig:iterativly_approach}
\end{figure}

\begin{table*}[t]
    \footnotesize
	\centering
	\caption{Datasets with their total number of samples, number of abnormal samples in the ground truth, and number of samples in $\mathcal{U}$ for different time windows ($\delta$).}
	\begin{tabular}{p{2.1cm}p{1.5cm}p{2.4cm} p{1.8cm}p{1.8cm}p{1.8cm}p{1.8cm}}
		\toprule
        \multirow{2}{*}{Dataset}  & \multirow{2}{*}{Total} & Abnormal  & \multicolumn{4}{c}{$\mathcal{U}$ for different $\delta$} \\
        \cmidrule{4-7}
                &  & (ground truth) & $\pm 1\,000\,ms$ & $\pm 5\,000\,ms$ & $\pm 10\,000\,ms$ & $\pm20\,000\,ms$ \\
        \midrule
        BGL           & 4\,747\,963 & 348\,460  & 392\,485    & 439\,656    & 459\,098    & 479\,030\\
        Thunderbird   & 5\,000\,000 & 226\,287  & 1\,423\,207 & 2\,368\,016 & 2\,853\,171 & 2\,924\,896\\
        Spirit        & 5\,000\,000 & 764\,890  & 1\,006\,419 & 2\,333\,800 & 3\,251\,475 & 3\,280\,803\\
        \bottomrule
	\end{tabular}
	\label{table:datasets}
\end{table*}

We propose an iterative training strategy to improve the anomaly detection capabilities of our method.
Suppose that in the first iteration, we train our model on the available data and obtain an anomaly score $\lVert z_i\rVert$ for each original log event $l_i \in \mathcal{L}$.
Instead of mapping the anomaly scores directly to labels based on a threshold, we instantiate a new model and use them as training input. Before, we smooth the anomaly scores to eliminate previously learned biases.
We assume that at least half of all samples are normal since anomalies are by definition~\cite{liang2007failure} comparably rare. Hence we calculate the median $m$ of all anomaly scores and subtract it from each anomaly score $\lVert z_i\rVert$ to obtain a smoothed anomaly score.
In the next step, we replace all negative scores with 0 to ensure that each score lives in the range $[0, \infty)$.
Lastly, we employ the hyperbolic tangent as a smoothing function, which effectively squeezes all anomaly scores to the range $[0, 1)$ and thus diminishes the influence of very large anomaly scores during the next training iteration. Eventually, this yields:

\begin{equation}
\label{eq:smooth}
\forall l_i \in \mathcal{L}: \tanh(\max(0, \lVert z_i\rVert - m))
\end{equation}

Following our problem definition, we then utilize the transformed anomaly scores as inaccurate labels of our log messages and train a new model.

\autoref{fig:iterativly_approach} depicts how the training labels are changing from iteration to iteration. 
Therefore the real anomaly is marked in dark red. The preliminary training label for this sample denotes an anomaly as well, as the respective log message is, along with other log messages, within the failure time window. 
The anomaly scores from the first iteration are then smoothed by utilizing \autoref{eq:smooth}. 
It can be observed that after the second iteration, the sample that was originally mislabeled receives a lower anomaly score than after the first iteration, and is from then on treated as normal.
This exemplifies the optimal process and the potential of our iterative method.

Due to omitting the application of a threshold in the intermediate iterations, both terms of our utilized loss function are enabled, and the prediction certainty of a model is effectively incorporated into the optimization procedure. 
After a configurable number of iterations, the final anomaly scores are eventually mapped to labels using a determined threshold.

\section{Evaluation}
\label{sec:evaluation}

This section evaluates PULL next to ten common methods from log anomaly detection and text classification.
We compare their performance on three different data sets, considering four different time windows $\delta$.

\subsection{Experimental Setup}

We first explain our experimental setup, namely which benchmark methods and datasets we chose and how we constructed our evaluation datasets.

\textbf{Datasets.} We evaluate PULL on three well known log datasets with anomalies that were recorded at different large-scale computer systems and presented in~\cite{oliner2007datasets}.
All datasets were labeled manually by experts.
The \emph{BGL} dataset is collected from a BlueGene/L supercomputer system at Lawrence Livermore National Labs (LLNL) and originates from a period of ~214 days, with on average 0.25 log messages per second.
The \emph{Spirit} dataset is collected from a Spirit supercomputer system at Sandia National Labs (SNL). 
We selected the first 5\,000\,000 log messages, that cover a period of ~48 days, with on average 1.2 log messages per second.
The \emph{Thunderbird} dataset is collected from a Thunderbird supercomputer system at SNL. 
Again, we selected the first 5\,000\,000 log messages that reflect a period of ~9 days, with on average 6.4 log messages per second.

We create our datasets by including all abnormal log events as well as their surrounding events within a time window $2*\delta$ in $\mathcal{U}$; all remaining log events are in $\mathcal{P}$. 
As different systems with different quality of monitoring solutions can only provide failure time windows with varying degrees of accuracy, we investigate the performance at four different time windows: $\pm 1000\,ms$ (2s), $\pm 5000\,ms$ (10s), $\pm 10000\,ms$ (20s), and $\pm 20000\,ms$ (40s). 
\autoref{table:datasets} presents the number of samples in $\mathcal{U}$ for each datasets.
For small failure time windows, class $\mathcal{U}$ does not deviate too much from the ground truth at BGL and Spirit. 
Consequently, these evaluation datasets contain only very few incorrectly labeled normal log messages.
For Thunderbird, on the other hand, $\mathcal{U}$ is already about six times larger than the ground truth at $\delta=1000\,ms$, meaning it is the hardest dataset for this task.

For larger $\delta$, $\mathcal{U}$ grows rapidly in all datasets until a certain limit.
At the largest observed windows, between $\delta=10000$ and $\delta=20000$, only relatively few new samples are added to $\mathcal{U}$.
This is reasonable, because systems often produce significantly more logs when an error occurs, placing them in close temporal proximity. 

\textbf{Benchmark Methods.} To obtain a significant and wide benchmark, we compare PULL to ten state-of-the-art text-classification and anomaly detection methods presented in a recent text-classification survey~\cite{kowsari2019text} as well as in an established survey for anomaly detection in system logs~\cite{he2016experience}.
Namely, we choose Deeplog, LogRobust, Rocchio, Invariants Miner, SVM, Boosting, PCA, Logistic Regression, Decision Tree, and Random Forest as benchmarks.
Many of these methods are explained in more detail in~\autoref{sec:related-work}.
DeepLog, Invariants Miner, and PCA are only able to train on samples in $\mathcal{P}$ to then detect deviations, while all others train on all samples in $\mathcal{P}$ and $\mathcal{U}$.

\textbf{Preprocessing and Model Setup.} The tokenization process as described in \autoref{sec:approach} is applied for all methods to ensure a fair comparison.
Each sequence of tokens $t_i$ is truncated to have a length of 26 for \emph{Thunderbird}, 18 for \emph{Spirit}, and 12 for \emph{BGL}. 
The dimensionality $d$ of our embeddings is set to 128.
We use Xavier weight initialization initially for model weights and embeddings.
For the training of our PULL model, we use a hidden dimensionality of 256, a batch size of 1024, a total of 8 epochs, and a dropout rate of 10\%. 
For optimization, we use the Adam optimizer with a learning rate of $10^{-4}$ and weight decay of $5 \cdot 10^{-5}$.

\subsection{Effect of Iterative Training}

\begin{figure}
    \centering
    \includegraphics[width=0.8\columnwidth]{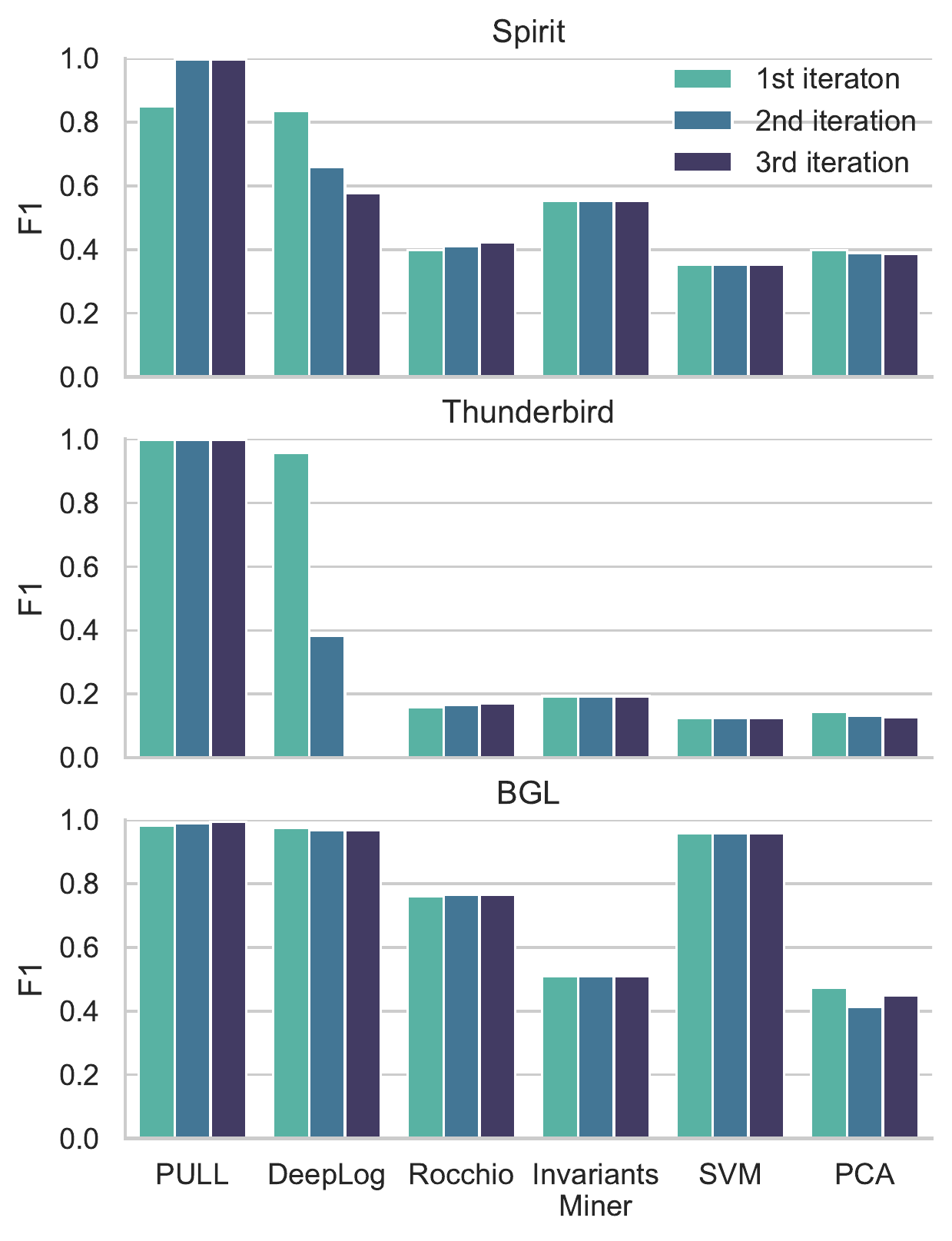}
    \caption{Performance after iterative training for $\delta=\pm 10000ms$. While PULL and Rocchio improve on each iteration, other methods degrade (e.g. DeepLog and PCA) or stay the same (e.g. Invariant Miners and SVM).}
    \label{fig:iterative_evaluation}
\end{figure}

\begin{figure*}
    \centering
    \includegraphics[width=0.9\textwidth]{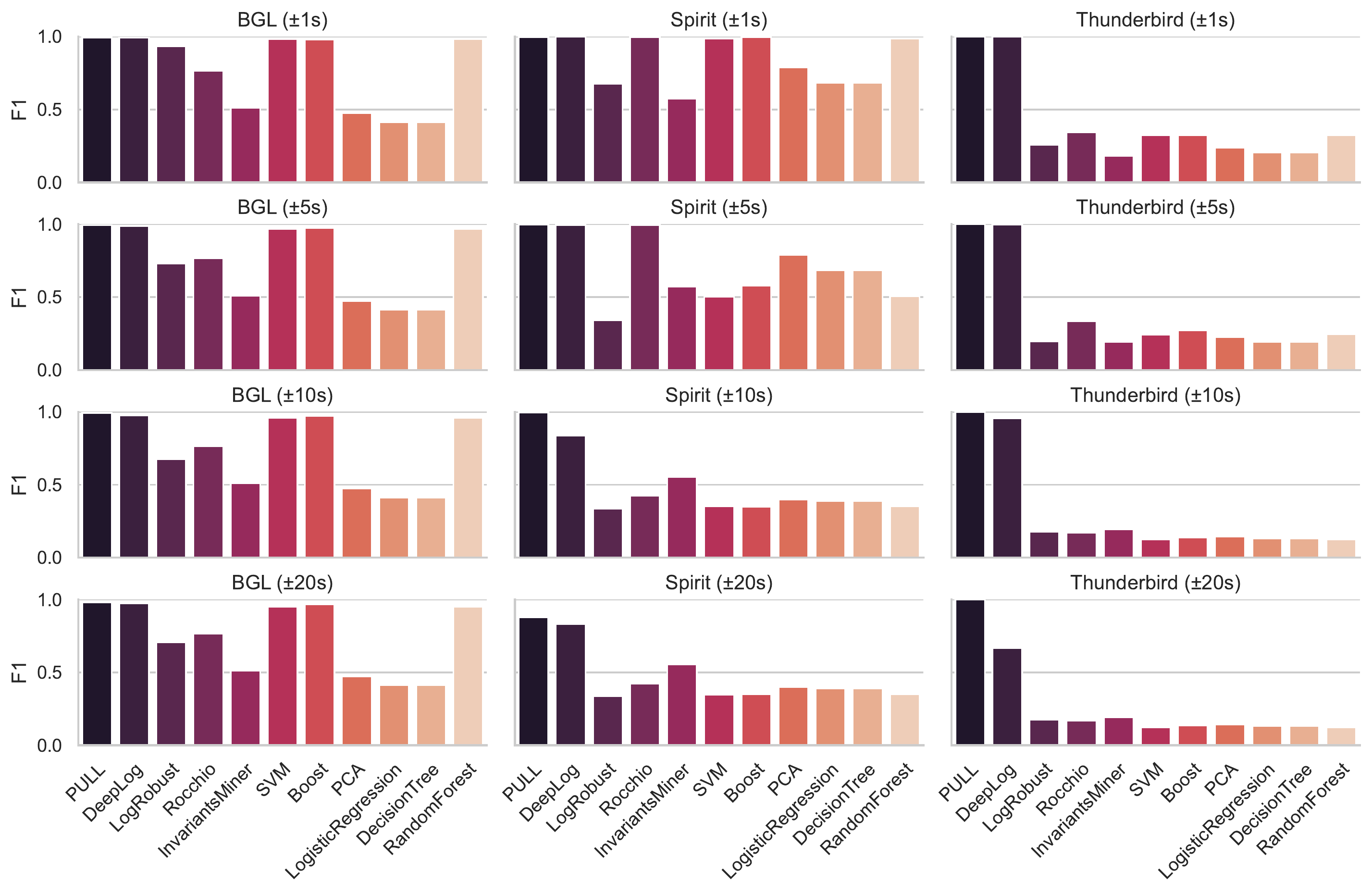}
    \caption{F1-scores of all experiments. PULL performs best across all evaluated datasets and time windows.}
    \label{fig:result}
\end{figure*}

Before comparing the performance of PULL to the benchmarks, we analyze the effect of iterative PU learning on all available methods to find out which methods benefit from this technique.
For this, we trained all methods three times on their respective output and observed the model performance after every iteration.
We chose three iterations to detect general trends for each method - in practice, the number of iterations would be determined by a suitable stopping condition or constraint.
\autoref{fig:iterative_evaluation} displays the results of some exemplary methods after every iteration.

The performance improvement or degradation was consistent across the three datasets for every method and can be classified into three categories: 

\begin{itemize}
    \item \emph{Performance Improvement.} Besides PULL, also the Roccio algorithm's performance improved consistently and often significantly after every iteration of training.
    For both methods, the range of possible anomalies is becoming smaller in each iteration, as more normal log messages are now labeled as such. 
    Consequently, there are now more actually anomalous messages in $\mathcal{U}$.
    Similarly, also the Random Forest's performance increased on every iteration, albeit only at the 4th digit behind the comma.
    \item \emph{Performance Degradation.} DeepLog and PCA's performance degraded after retraining, meaning that an increasing number of actual anomalies are classified as normal by these methods. 
    As a result, anomalies are now considered normal during further iterations, whereby the model tends to classify similar anomalies as normal as well. 
    \item \emph{No Change in Performance.} LogRobust, Decision Tree, Logistic Regression, and Invariants Miner did not change their performance at all between rounds of training as these methods have obtained their optimal result already during the first iteration. 
    The boosting method and the SVM did not change their performance either, albeit minor fluctuations in the precision and recall at 4 or 5 digits behind the comma.
\end{itemize}

\subsection{Comparison of Methods}

We now assess the performance of PULL and all baselines in terms of precision, recall, and F1-score metrics.
We depict the best F1-scores for each method in \autoref{fig:result}.
For methods that benefit from the iterative approach (PULL, Rocchio, Random Forest), the F1-scores after the third iteration are displayed. For the remaining methods, we display the F1-score after a single iteration of training. 
PULL achieves the highest F1-score across all experiments.
As expected, with increasing $\delta$ and thus growing size of $\mathcal{U}$, the performance across all methods tends to decrease.

For $\delta = \pm 1000\,ms$, most methods achieve good performance. An exception is the Thunderbird dataset, which is characterized by a large $\mathcal{U}$ class, even for $\delta \pm 1000$: No methods but PULL and DeepLog manage to achieve an F1-score higher than 0.35, while PULL reaches 0.999 precision at perfect recall and DeepLog reaches 0.925 precision at nearly perfect recall.

For $\delta = \pm 5000\,ms$, the performance of all methods but PULL, DeepLog, and Rocchio now also degrades on the Spirit dataset.
We can observe that simple methods tend to have a close-to-perfect recall 
but suffer from bad precision, meaning that they fail to identify normal log messages in $\mathcal{U}$. %
In contrast to Thunderbird and Spirit, most methods 
continue to achieve very good results on the BGL dataset.

The biggest drop in performance becomes evident at the time window of $\delta = \pm 10000\,ms$. For the BGL dataset, we notice a notable drop in F1-scores of other methods to 0.97, while only PULL maintains its high performance of more than 0.994. 
On the Spirit dataset, the F1-score of most methods drops below 0.5 while PULL still reaches 0.997. 
For Thunderbird, the precision of other methods drops to only 0.1, meaning that only one out of 10 log events labeled as abnormal are actually abnormal. 

Although, as explained above, the size of $\mathcal{U}$ does not change significantly when increasing the window from $\delta = \pm 10000\,ms$ to $\delta = \pm 20000\,ms$, we notice some notable drops in performance. 
Specifically, PULL now only reaches an F1-score of 0.87 on the Spirit dataset, followed by DeepLog with an F1-Score of 0.83. On Thunderbird, DeepLog's performance decreases to 0.67 while PULL maintains its F1-Score of 0.9995 from the previous time window.

Overall, it can be observed that PULL and DeepLog, which are both based on neural networks, perform significantly better than most traditional methods.
An exception to this is LogRobust, which is also based on a neural network, but suffers from overfitting and is not able to reclassify incorrectly labeled input data. Therefore, LogRobust is not applicable for a PU learning strategy, which demonstrates that neural networks are not generally superior.
While traditional methods try to distinguish between black and white, some neural network based methods get a better intuition of what constitutes an anomaly.

Nevertheless, some threats to validity remain. On the one hand, we make the assumption that at least half of the data is normal, however, in practice it can be assumed that the majority of data is normal and only few anomalies exist.
On the other hand, the failure times must be roughly known, which, however, is not as strong a precondition as labeled data, since they can be deduced from commonly employed monitoring solutions whereas labeling requires manual effort.

\section{Conclusion}
\label{sec:conclusion}
This paper presents PULL, a weakly supervised model for reactive anomaly detection in log data.
PULL requires rough estimates of failure time windows provided by monitoring systems and identifies anomalous log messages within these windows with very high precision and recall, which qualifies it for application in real-world use cases.
Therefore, labeled training data, which is usually difficult to obtain due to the required need for time-consuming manual labeling through experts, can be omitted.
PULL is based on the attention mechanism and uses a custom objective function for weak supervision deep learning techniques that accounts for imbalanced data and deals with inaccurate labels.
We further propose and investigate a novel training strategy that iteratively trains a newly instantiated model on the regulated output of previous models, thereby improving the performance of PULL specifically while leaving most baselines unchanged.

We evaluated PULL in comparison to ten benchmark methods on three common datasets and at four different failure time windows.
It outperforms other methods across all experiments and reaches a performance of more than $0.994$ F1-score, even for failure time windows as large as $\pm 10000\,ms$.%

Regarding future work, we want to incorporate additional data sources to obtain other weakly and inaccurately labeled data for the training process. 
Moreover, we want to further develop our method such that it does not only localize the anomalies but also identifies the underlying root causes.

\bibliographystyle{ieeetr}
\bibliography{references}

\begin{thebibliography}{10}

\bibitem{rosendo2018improve}
D.~Rosendo, G.~Leoni, D.~Gomes, A.~Moreira, G.~Gon{\c{c}}alves, P.~Endo,
  J.~Kelner, D.~Sadok, and M.~Mahloo, ``How to improve cloud services
  availability? investigating the impact of power and it subsystems failures,''
  in {\em HICSS}, 2018.

\bibitem{gulenko2020ai}
A.~Gulenko, A.~Acker, O.~Kao, and F.~Liu, ``Ai-governance and levels of
  automation for aiops-supported system administration,'' in {\em ICCCN}, IEEE,
  2020.

\bibitem{sridharan2018distributed}
C.~Sridharan, {\em Distributed systems observability: a guide to building
  robust systems}.
\newblock O'Reilly Media, 2018.

\bibitem{wittkopp2021taxonomy}
T.~Wittkopp, P.~Wiesner, D.~Scheinert, and O.~Kao, ``A taxonomy of anomalies in
  log data,'' in {\em 2nd International Workshop on Artificial Intelligence for
  IT Operations (AIOPS)}, 2021.

\bibitem{du2017deeplog}
M.~Du, F.~Li, G.~Zheng, and V.~Srikumar, ``Deeplog: Anomaly detection and
  diagnosis from system logs through deep learning,'' in {\em SIGSAC}, 2017.

\bibitem{zhang2019robust}
X.~Zhang, Y.~Xu, Q.~Lin, B.~Qiao, H.~Zhang, Y.~Dang, C.~Xie, X.~Yang, Q.~Cheng,
  Z.~Li, {\em et~al.}, ``Robust log-based anomaly detection on unstable log
  data,'' in {\em ESEC/FSE}, 2019.

\bibitem{Wittkopp_LogLab_2021}
T.~Wittkopp, P.~Wiesner, D.~Scheinert, and A.~Acker, ``Loglab: Attention-based
  labeling of log data anomalies via weak supervision,'' in {\em International
  Conference on Service-Oriented Computing (ICSOC)}, 2021.

\bibitem{nedelkoski2020self}
S.~Nedelkoski, J.~Bogatinovski, A.~Acker, J.~Cardoso, and O.~Kao,
  ``Self-attentive classification-based anomaly detection in unstructured
  logs,'' in {\em ICDM}, IEEE, 2020.

\bibitem{wittkopp2021a2log}
T.~Wittkopp, A.~Acker, S.~Nedelkoski, J.~Bogatinovski, D.~Scheinert, W.~Fan,
  and O.~Kao, ``A2log: Attentive augmented log anomaly detection,'' in {\em
  HICSS}, 2022.

\bibitem{yang2021semi}
L.~Yang, J.~Chen, Z.~Wang, W.~Wang, J.~Jiang, X.~Dong, and W.~Zhang,
  ``Semi-supervised log-based anomaly detection via probabilistic label
  estimation,'' in {\em ICSE}, IEEE, 2021.

\bibitem{wen2021time}
Q.~Wen, L.~Sun, F.~Yang, X.~Song, J.~Gao, X.~Wang, and H.~Xu, ``Time series
  data augmentation for deep learning: {A} survey,'' in {\em IJCAI}, ijcai.org,
  2021.

\bibitem{sukhwani2017monitoring}
H.~Sukhwani, R.~Matias, K.~S. Trivedi, and A.~Rindos, ``Monitoring and
  mitigating software aging on ibm cloud controller system,'' in {\em ISSREW},
  IEEE, 2017.

\bibitem{liu2002partially}
B.~Liu, W.~S. Lee, P.~S. Yu, and X.~Li, ``Partially supervised classification
  of text documents,'' in {\em ICML}, 2002.

\bibitem{liu2003building}
B.~Liu, Y.~Dai, X.~Li, W.~S. Lee, and P.~S. Yu, ``Building text classifiers
  using positive and unlabeled examples,'' in {\em ICDM}, IEEE, 2003.

\bibitem{DevlinCLT19}
J.~Devlin, M.~Chang, K.~Lee, and K.~Toutanova, ``{BERT:} pre-training of deep
  bidirectional transformers for language understanding,'' in {\em NAACL-HLT},
  Association for Computational Linguistics, 2019.

\bibitem{VaswaniSPUJGKP17}
A.~Vaswani, N.~Shazeer, N.~Parmar, J.~Uszkoreit, L.~Jones, A.~N. Gomez,
  L.~Kaiser, and I.~Polosukhin, ``Attention is all you need,'' in {\em
  NeurIPS}, 2017.

\bibitem{zhou2018brief}
Z.-H. Zhou, ``A brief introduction to weakly supervised learning,'' {\em
  National science review}, 2018.

\bibitem{zhu2009introduction}
X.~Zhu and A.~B. Goldberg, ``Introduction to semi-supervised learning,'' {\em
  Synthesis lectures on artificial intelligence and machine learning}, 2009.

\bibitem{bekker2020learning}
J.~Bekker and J.~Davis, ``Learning from positive and unlabeled data: A
  survey,'' {\em Machine Learning}, 2020.

\bibitem{fusilier2015detecting}
D.~H. Fusilier, M.~Montes-y G{\'o}mez, P.~Rosso, and R.~G. Cabrera, ``Detecting
  positive and negative deceptive opinions using pu-learning,'' {\em
  Information processing \& management}, 2015.

\bibitem{WuCWWZW20}
Z.~Wu, J.~Cao, Y.~Wang, Y.~Wang, L.~Zhang, and J.~Wu, ``hpsd: {A} hybrid
  pu-learning-based spammer detection model for product reviews,'' {\em {IEEE}
  Trans. Cybern.}, 2020.

\bibitem{LuoCLJ18}
Y.~Luo, S.~Cheng, C.~Liu, and F.~Jiang, ``{PU} learning in payload-based web
  anomaly detection,'' in {\em SSIC}, {IEEE}, 2018.

\bibitem{mordelet2014bagging}
F.~Mordelet and J.-P. Vert, ``A bagging svm to learn from positive and
  unlabeled examples,'' {\em Pattern Recognition Letters}, 2014.

\bibitem{he2016experience}
S.~He, J.~Zhu, P.~He, and M.~R. Lyu, ``Experience report: System log analysis
  for anomaly detection,'' in {\em ISSRE}, IEEE, 2016.

\bibitem{kowsari2019text}
K.~Kowsari, K.~Jafari~Meimandi, M.~Heidarysafa, S.~Mendu, L.~Barnes, and
  D.~Brown, ``Text classification algorithms: A survey,'' {\em Information},
  2019.

\bibitem{hosmer2013applied}
D.~W. Hosmer~Jr, S.~Lemeshow, and R.~X. Sturdivant, {\em Applied logistic
  regression}.
\newblock John Wiley \& Sons, 2013.

\bibitem{genkin2007large}
A.~Genkin, D.~D. Lewis, and D.~Madigan, ``Large-scale bayesian logistic
  regression for text categorization,'' {\em technometrics}, 2007.

\bibitem{quinlan1986induction}
J.~R. Quinlan, ``Induction of decision trees,'' {\em Machine learning}, 1986.

\bibitem{safavian1991survey}
S.~R. Safavian and D.~Landgrebe, ``A survey of decision tree classifier
  methodology,'' {\em IEEE transactions on systems, man, and cybernetics},
  1991.

\bibitem{chen2004failure}
M.~Chen, A.~X. Zheng, J.~Lloyd, M.~I. Jordan, and E.~Brewer, ``Failure
  diagnosis using decision trees,'' in {\em ICAC}, IEEE, 2004.

\bibitem{ho1995random}
T.~K. Ho, ``Random decision forests,'' in {\em ICDAR}, IEEE, 1995.

\bibitem{manevitz2001one}
L.~M. Manevitz and M.~Yousef, ``One-class svms for document classification,''
  {\em Journal of machine Learning research}, 2001.

\bibitem{liang2007failure}
Y.~Liang, Y.~Zhang, H.~Xiong, and R.~Sahoo, ``Failure prediction in ibm
  bluegene/l event logs,'' in {\em ICDM}, IEEE, 2007.

\bibitem{sowmya2016large}
B.~Sowmya, K.~Srinivasa, {\em et~al.}, ``Large scale multi-label text
  classification of a hierarchical dataset using rocchio algorithm,'' in {\em
  CSITSS}, IEEE, 2016.

\bibitem{selvi2017text}
S.~T. Selvi, P.~Karthikeyan, A.~Vincent, V.~Abinaya, G.~Neeraja, and
  R.~Deepika, ``Text categorization using rocchio algorithm and random forest
  algorithm,'' in {\em ICoAC}, IEEE, 2017.

\bibitem{LiCJHY20}
X.~Li, P.~Chen, L.~Jing, Z.~He, and G.~Yu, ``Swisslog: Robust and unified deep
  learning based log anomaly detection for diverse faults,'' in {\em ISSRE},
  {IEEE}, 2020.

\bibitem{GuoYW21}
H.~Guo, S.~Yuan, and X.~Wu, ``Logbert: Log anomaly detection via {BERT},'' in
  {\em IJCNN}, {IEEE}, 2021.

\bibitem{YangCWWJDZ21}
L.~Yang, J.~Chen, Z.~Wang, W.~Wang, J.~Jiang, X.~Dong, and W.~Zhang,
  ``Semi-supervised log-based anomaly detection via probabilistic label
  estimation,'' in {\em ICSE}, {IEEE}, 2021.

\bibitem{jolliffe2005principal}
I.~Jolliffe, ``Principal component analysis,'' {\em Encyclopedia of statistics
  in behavioral science}, 2005.

\bibitem{lou2010mining}
J.~Lou, Q.~Fu, S.~Yang, Y.~Xu, and J.~Li, ``Mining invariants from console logs
  for system problem detection,'' in {\em ATC}, USENIX, 2010.

\bibitem{schapire2000boostexter}
R.~E. Schapire and Y.~Singer, ``Boostexter: A boosting-based system for text
  categorization,'' {\em Machine learning}, 2000.

\bibitem{LinZLZC16}
Q.~Lin, H.~Zhang, J.~Lou, Y.~Zhang, and X.~Chen, ``Log clustering based problem
  identification for online service systems,'' in {\em ICSE}, {ACM}, 2016.

\bibitem{he2017drain}
P.~He, J.~Zhu, Z.~Zheng, and M.~R. Lyu, ``Drain: An online log parsing approach
  with fixed depth tree,'' in {\em ICWS}, IEEE, 2017.

\bibitem{MengLZZPLCZTSZ19}
W.~Meng, Y.~Liu, Y.~Zhu, S.~Zhang, D.~Pei, Y.~Liu, Y.~Chen, R.~Zhang, S.~Tao,
  P.~Sun, and R.~Zhou, ``Loganomaly: Unsupervised detection of sequential and
  quantitative anomalies in unstructured logs,'' in {\em IJCAI}, ijcai.org,
  2019.

\bibitem{FarzadG20}
A.~Farzad and T.~A. Gulliver, ``Unsupervised log message anomaly detection,''
  {\em {ICT} Express}, 2020.

\bibitem{YuanALYL020}
Y.~Yuan, S.~S. Adhatarao, M.~Lin, Y.~Yuan, Z.~Liu, and X.~Fu, ``{ADA:} adaptive
  deep log anomaly detector,'' in {\em INFOCOM}, {IEEE}, 2020.

\bibitem{GehringAGYD17}
J.~Gehring, M.~Auli, D.~Grangier, D.~Yarats, and Y.~N. Dauphin, ``Convolutional
  sequence to sequence learning,'' in {\em ICML}, {PMLR}, 2017.

\bibitem{oliner2007datasets}
A.~Oliner and J.~Stearley, ``What supercomputers say: A study of five system
  logs,'' in {\em DSN}, 2007.

\end{thebibliography}

\balance

\end{document}